\documentclass[USLetter,conference]{IEEEtran}
\usepackage{cite}
\usepackage{amsmath,graphicx,amssymb,siunitx}
\usepackage{float,dblfloatfix,subfig}
\usepackage{verbatim}
\usepackage{tabularx}
\usepackage{mathabx}
\usepackage{mwe,tikz}
\usetikzlibrary{spy}
\ifCLASSINFOpdf
\else
\fi

\begin{document}
\title{Realistic Hair Simulation Using Image Blending}
\author{
\IEEEauthorblockN{Mohamed~Attia\IEEEauthorrefmark{1}, Mohammed~Hossny\IEEEauthorrefmark{1}, Saeid~Nahavandi\IEEEauthorrefmark{1},
Anousha Yazdabadi\IEEEauthorrefmark{2} and
Hamed Asadi\IEEEauthorrefmark{2}}
\IEEEauthorblockA{\IEEEauthorrefmark{1}Institute for Intelligent Systems Research and Innovation (IISRI), Deakin University}\IEEEauthorrefmark{2}School of Medicine, Deakin University}

\maketitle

\begin{abstract}
In this presented work, we propose a realistic hair simulator using image blending for dermoscopic images. This hair simulator can be used for benchmarking and validation of the hair removal methods and in data augmentation for improving computer aided diagnostic tools. We adopted one of the popular implementation of image blending to superimpose realistic hair masks to hair lesion. This method was able to produce realistic hair masks according to a predefined mask for hair. Thus, the produced hair images and masks can be used as ground truth for hair segmentation and removal methods by inpainting hair according to  a pre-defined hair masks on hair-free areas. Also, we achieved a realism score equals to 1.65 in comparison to 1.59 for the state-of-the-art hair simulator.
\end{abstract}

\IEEEpeerreviewmaketitle

\section{Introduction}
\begin{figure}[t]
    \centering
    \includegraphics[width=1\linewidth]{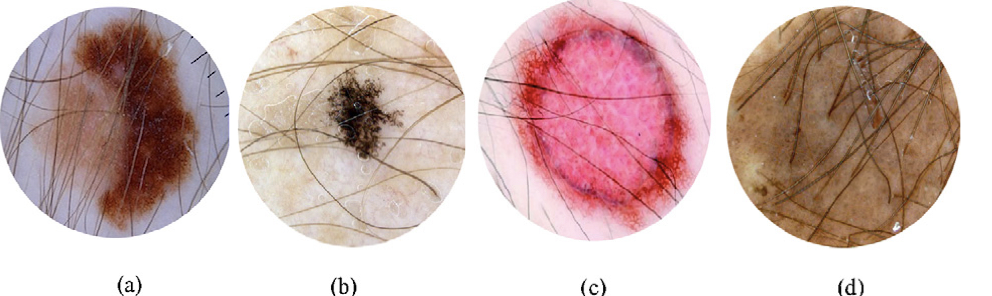}
    \caption{Skin images are usually occluded by hair artefacts. These artefacts affects the accuracy of computer aided diagnostic tools.}
    \label{fig1:skinimgs}
\end{figure}

Automated hair removal is one of the most important application for recent advances for artificial intelligence and deep learning~\cite{esteva2017dermatologist}. This is because hair artefacts affect automated skin cancer diagnosis tools. These tools include lesion segmentation and classification into malignant and benign~\cite{attia2016skin,attia2017skin}. However, lesions are usually occluded by hair and other artefacts, as shown in Fig.~\ref{fig1:skinimgs}. Most importantly, the wide adoption of deep learning in computer vision recent tasks in general and medical imaging in particular, has dictated preprocessing artefact removal steps~\cite{Abobakr2016CNN,abobakr2017rula,abobakr2017kinect,Khaled2016}. Thus, hair removal is a crucial part of the automated analysis to improve the classification accuracy and enhance the ability of the models to generalise~\cite{yuan2017automatic}. Many digital hair removal methods were proposed to remove the super imposed hair in a non-invasive fashion. They are based on two main steps: hair segmentation and hair removal~\cite{koehoorn2016effcient}. These methods were able to deal with different hair structures. However, they have been always validated on relatively small datasets, that have many properties in common in terms of size, colours, illuminations. Thus, these methods have limited capability to generalise to larger number of images and the code is not publicly available for comparison.

\smallbreak
Several studies tried to overcome these problems by simulating hair by black curved lines with random length and curvatures. However, the resultant hair was unrealistic with solid black colours. To ensure high degree of realism, the simulated hair shall be having realistic curvature and colours. In this presented work, we use image blending techniques using poison blending to blend real hair masks into lesion to  simulate hair occlusion on lesions. We blended a pre-segmented hair masks with hair-free image. This methods was able to achieve more realistic results with high degree of variability. Image blending techniques aims at blending image with background to achieve high compatibility between the source and destination. One of the main advantages of image blending is the blended image depends on both the source and destination. Thus, for every pair of images, the hair shapes and colours differ. 
\smallbreak
Quantitative analysis of the proposed hair simulator is one the most challenging parts. Most of the recent works in the hair simulation was proposed based on the qualitative analysis without comparison to real images. The lack of the quantitative metrics for image quality will affect the applicability of parametric analysis of the method or benchmarking. However, recent advances in blind image quality metric facilitates the characterisation of the natural images based on morphology and colour compatibility. These methods have been validated against human observer. 
\smallbreak
The main contribution of this presented work is based on the novel proposed methodology for realistic hair simulation using image blending techniques as a tool for validation of hair segmentation and inpainting methods. Also, we used blinded image quality tool for quantitative analysis of the proposed method. The rest of this  paper is organised as follows. Section~\ref{sec:related} describes the efforts documented in the literature for state-of-the-art methods. In Section~\ref{sec:METHODS}, a brief description of the proposed methodology is elaborated. followed by the discussion for the obtained results of the conducted experiments. In Section~\ref{sec:CONCLUSION}, conclusions are drawn from the implementation of the proposed architecture.

\section{Related Work}
\label{sec:related}
To solve hair the occlusion problem, many methods have been introduced. As shown in Table~\ref{table:comparison}, they vary between each others in terms of the assumptions of hair morphology and the hair detection method. Some of these methods were based on morphological operations for hair detection~\cite{lee1997dullrazor,Zhou:2014:IIC:2683354.2683359}. On the other hand, some are based on matched filters~\cite{abbas2011unsupervised,abbas2011hair}.
\smallbreak
Koehoorn~\textit{et al.} proposed a comprehensive framework for segmentation of hair with different morphology and colours~\cite{koehoorn2016effcient}. They segmented the hair based on multi-scale skeletonization that has been able to detect hair at different scales including, short, long, straight and curly hair. They compared their proposed method against other different hair detection methods using over than 300 manually annotated images. In these proposed methods, they validated their methods on relatively small dataset with size near 300 images. This is due to the difficulty of manual annotation of the data. Also, the comparison was not based on a publicly available datsets which limits the ability to compare or validate the results. 
\smallbreak
To partially solve the limited data problem, several attempts for hair synthesis was proposed. In these methods, they made several assumption for hair synthesis. They simulated the hair according to the external morphology of the hair based on the assumptions: the hair has thin architecture with varying width. Nevertheless, they based their simulation techniques on utilisation of simplified mathematical models, along with other assumption, to synthesise skin hair. Therefore, the synthesised hair suffers from several artefacts.
\smallbreak
Denton~\textit{et al.} proposed one of the earliest attempts to simulate hair  for skin lesion~\cite{denton1997synthesis}. She~\textit{et al.} proposed used the hair morphology to design a hair simulator that produces a single black hair image. This hair is 100 pixels length and 3 pixels wide. They limited hair orientations at horizontal, vertical and 45$\deg$~\cite{she2001improved}. Then, they improved the skin hair simulator and they introduced it as a part of skin lesion simulator to facilitate the analysis of the skin lesions. They simulated the hair as straight and curved black lines~\cite{she2006simulation}. However, the results were based on fixed color hair simulation.

\begin{figure}[!t]
    \centering
\begin{tabular}{ccc}
    \includegraphics[width=0.3\linewidth]{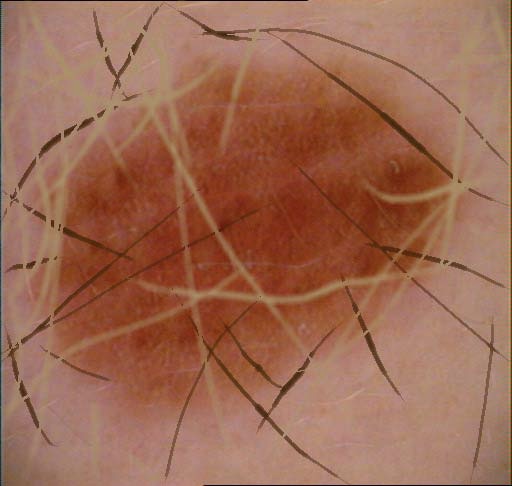} & \includegraphics[width=0.3\linewidth]{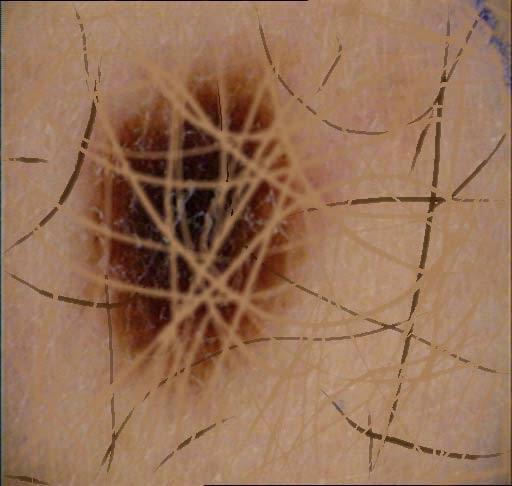} & \includegraphics[width=0.3\linewidth]{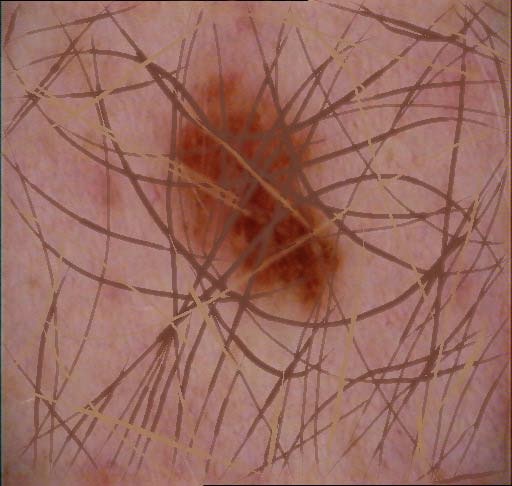}
\end{tabular}    \caption{The simulated hair using the method proposed by Mirzaalian~\textit{et al.}~\cite{mirzaalian2014hair}. The simulated images suffer from unrealistic hair structure with incompatible colour to the skin lesion colours.. }
    \label{fig2:denton}
\end{figure}

\smallbreak
Mirzaalian~\textit{et al.} proposed a hair simulator based on random generated curved lines. They prepared medial curves of the hairs generated by a random curve synthesizer. Then, they dilated the generated curves to achieve hair-thickened curve by dilation with a disk structuring element of varying radius to simulate average human hair. Then, they coloured the generated mask by using a preset dictionary. They limited the hair colours to which includes yellow, brown, white, black and grey. They applied gaussian filter on the edges to blend the hair with the background\cite{mirzaalian2014hair}. However, the hair colours and hair structure are not realistic as shown in Fig.~\ref{fig2:denton}.
\smallbreak
To address this problem, we are going to used pre-segmented hair masks and superimpose them on lesion images to blend them. Image blending has been used for data augmentation to produce images with the corresponding ground-truth~\cite{georgakis2017synthesizing,park2015articulated}. The main advantage of this method is achieving realistic results with expected variation in morphology and appearance, along with the ground truth~\cite{khoreva2017lucid,tang2013learning,duchateau2018model}.
\smallbreak
Many approached were proposed to assess the realism of the composite and synthesized, based on parametric metrics. They characterised the images based on noise level, blur level and color transformations. However, these methods are hard to generalise to include images from different sources. Zhu~\textit{et al.} proposed a non-parametric characterisation methodology for assessment of image realism~\cite{zhu2015learning}. They used a generizable model for discrimination of realistic and fake images. They trained a deep convolutional model based on VGG architecture to perform a non-task-specific discriminative classifier using end-to-end paradigm~\cite{liu2018normalized}.  

\begin{table*}[ht]
    \centering
	{

		\begin{tabular}{l|llcc}
			\textbf{Method} & \textbf{Hair detection} &\textbf{ Compared with} & \textbf{\#images} & \textbf{Code} \\
			\hline
			\hline~\\
			Dull razor~\cite{lee1997dullrazor}&
			Morphological closing &-& 5 & A\\~\\

			\hline~\\

			E-shaver~\cite{kiani2011shaver}&Prewitt edge detector  & Dull razor~\cite{lee1997dullrazor} &50&N/A  \\~\\

			\hline~\\

			Schmid\textit{et al.}~\cite{schmid2003towards}& Morphological closing& &200&N/A\\~\\

			\hline~\\

			Zhou~\textit{et al.}~\cite{zhou2008feature}&\begin{tabular}[t]{@{}l@{}}
			Line detection\\ Curve fitting
			\end{tabular}&Schmid\textit{et al. }~\cite{schmid2003towards} &460&N/A\\~\\

			\hline~\\

			Huang~\textit{et al.}~\cite{huang2013robust} &Multi-scale matched  filters  & Dull razor~\cite{lee1997dullrazor} &20  &N/A\\~\\

			\hline~\\

			Fiorese~\textit{et al.}~\cite{fiorese2011virtualshave}&  Top-Hat operator& Dull razor~\cite{lee1997dullrazor} &20  &N/A \\~\\

			\hline

			Xie~\textit{et al.}~\cite{Xie2013}& Top-Hat operator &  DullRazor~\cite{lee1997dullrazor}& 40  &
				N/A  \\~\\

			\hline~\\

			Abbas~\textit{et al.}~\cite{abbas2011hair}& Derivatives of Gaussian  & \begin{tabular}[t]{@{}l@{}}  Dull razor~\cite{lee1997dullrazor} \\ Xie~\textit{et al.}~\cite{xie2015no} \\ Zhou~\textit{et al.}~\cite{zhou2008feature}\end{tabular} & 100  &N/A  \\~\\

			\hline~\\

			Maglogiannis~\textit{et al.}~\cite{maglogiannis2015hair} &  \begin{tabular}[t]{@{}l@{}} Bottom-Hat transform \\ Laplacian \\ Laplacian of Gaussian\\ Sobel \end{tabular}  &  Dull razor~\cite{lee1997dullrazor}& 10 &N/A \\~\\

			\hline~\\

			Mirzaalian~\textit{et al.} ~\cite{mirzaalian2014hair}&Matched filters&-&\begin{tabular}[t]{@{}l@{}l@{}}40 real\\94 synthetic \end{tabular}&A\\~\\

			\hline~\\

			Du~\textit{et al.}~\cite{du2017hair}&   \begin{tabular}[t]{@{}l@{}l@{}}Top-Hat transform \\ Multi-scale curvilinear \\ Matched filtering \end{tabular}  & \begin{tabular}[t]{@{}l@{}l@{}l@{}l@{}l@{}} DullRazor~\cite{lee1997dullrazor} \\ Xie \textit{et al.}~\cite{xie2009pde} \\ Huang \textit{et al.}~\cite{huang2013robust}\\  Fiorese \textit{et al.}~\cite{fiorese2011virtualshave} \\ Abbas \textit{et al.}~\cite{abbas2011hair} \\ Virtual Shaver~\cite{koehoorn2015automated} \end{tabular} &  60 & N/A \\~\\

			\hline~\\

			Virtual shaver~\cite{koehoorn2015automated}&  \begin{tabular}[t]{@{}l@{}} Multiscale skeleton \\ Morphological operators \end{tabular} & \begin{tabular}[t]{@{}l@{}l@{}l@{}l@{}} Dull razor~\cite{lee1997dullrazor} \\ Xie~\textit{et al.}~\cite{xie2015no}  \\ Huang~\textit{et al.}~\cite{huang2013robust}\\  Fiorese~\textit{et al.}~\cite{fiorese2011virtualshave} \\ Abbas~\textit{et al.}~\cite{abbas2011hair}\end{tabular} & over 300 & A \\

		\end{tabular}
	}
	\caption{Summarization for segmentation methods in terms of hair detection method, comparison to other implementation, number of test images, availability of the implementation and the used color space. ``A'' stands for available code, while, ``N/A'' stands for not available code.}\label{table:comparison}
\end{table*}

\section{Proposed Method}
\label{sec:METHODS}
Image blending can be mathematically formulated as follows in Eqn.~\ref{Eqn:blending}.
\begin{equation}
    I_{blended}=I_{source}*M+I_{destination}* (1-M)
    \label{Eqn:blending}
\end{equation}
where $I_{blended}$ is the output image, $I_{source}$ is the hair mask source, $I_{destination}$ is the hair-free image and $M$ is the binary mask for hair location.
However, image blending faces two main challenges: the harmonisation of the image colour and the blending of the image borders ``colour bleeding''. These challenges can be addressed by applying additional constraints on the gradients along the boundaries of the region to be replaced $I_{source}*(1-M)$ to match the enclosing image $I_{destination}* (1-M)$~\cite{zhu2018comparative,wang2017avoiding}. Thus, the image editing must be done on the boundaries using a guidance vector field. This guidance vector field is estimated for each colour channel separately~\cite{perez2003poisson}.
\smallbreak
Perez~\textit{et al.} proposed an image blending algorithm based on Poisson equation. They used a large sparse linear system to find the optimal solution for this Poisson equation, along with the constraint on the vector guidance field, as shown in Eqn.~\ref{Eqn:poisson_original}~\cite{perez2003poisson}. 
\begin{equation}
\min_{r}\int\int|\nabla r|^{2}, \forall~r|_{\partial m}= d|_{\partial m}
\label{Eqn:poisson_original}    
\end{equation}
where $r$ is the output image, $r|\partial m$ is the boundary of the masked area to be blended, and $d|\partial m$ is boundary of the destination image.
Despite the robustness of the mathematically solution of the poisson equation, it works fine in the case of smooth change in the intensity across the boundary. Otherwise, the blended image suffers from bleeding artefacts~\cite{zhu2018comparative}. Dizdaroglu~\textit{et al.} proposed a method based on utilisation of colour information along with the gradients to reduce colour bleeding~\cite{dizdarouglu2011improved}.
\begin{equation}
\min_{r}\int\int|\nabla r - \nabla s|^{2}, \forall~r|_{\partial m}= d|_{\partial m}
\label{Eqn:poisson_modified}    
\end{equation}
where $r$ is the output image, $s$ is the source image, $r|\partial m$ is the boundary of the masked area to be blended, and $d|\partial m$ is boundary of the destination image.
Thus, Adding the gradient of the source image in the minimisation equation reduced the colour bleeding, as shown in Eqn.~\ref{Eqn:poisson_modified}. However, the output suffers from the domination of the source colours on the blended areas in the image. 
\smallbreak
Afifi~\textit{et al.} proposed a modified poison equation by minimising the scalar over not only the source but also the destination. They performed image blending on two steps: blending the source and the target according to the complement of the mask to produce an intermediate image. They used this intermediate image as source for blending with the target, according to the original mask. 
\smallbreak 
In our proposed method, we are going to adopt this modified version. This method suits the application, as it will change the hair morphology according to both of the source and destination. This will help in data augmentation process, where the hair will have different external appearance based on the combination of both the source and destination images.

\section{Experimental Results}
\begin{figure*}
    \centering
    \begin{tabular}{ccc}
    Image&Mask&Simulated Image\\
    \includegraphics[width=0.3\linewidth]{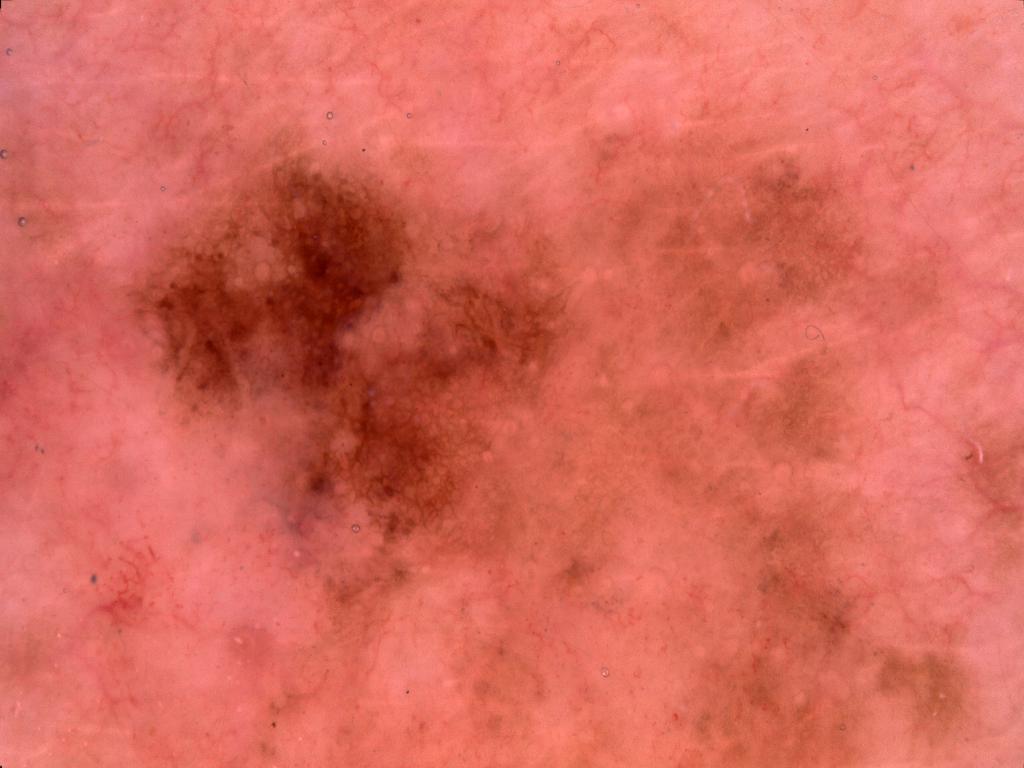}&\includegraphics[width=0.3\linewidth]{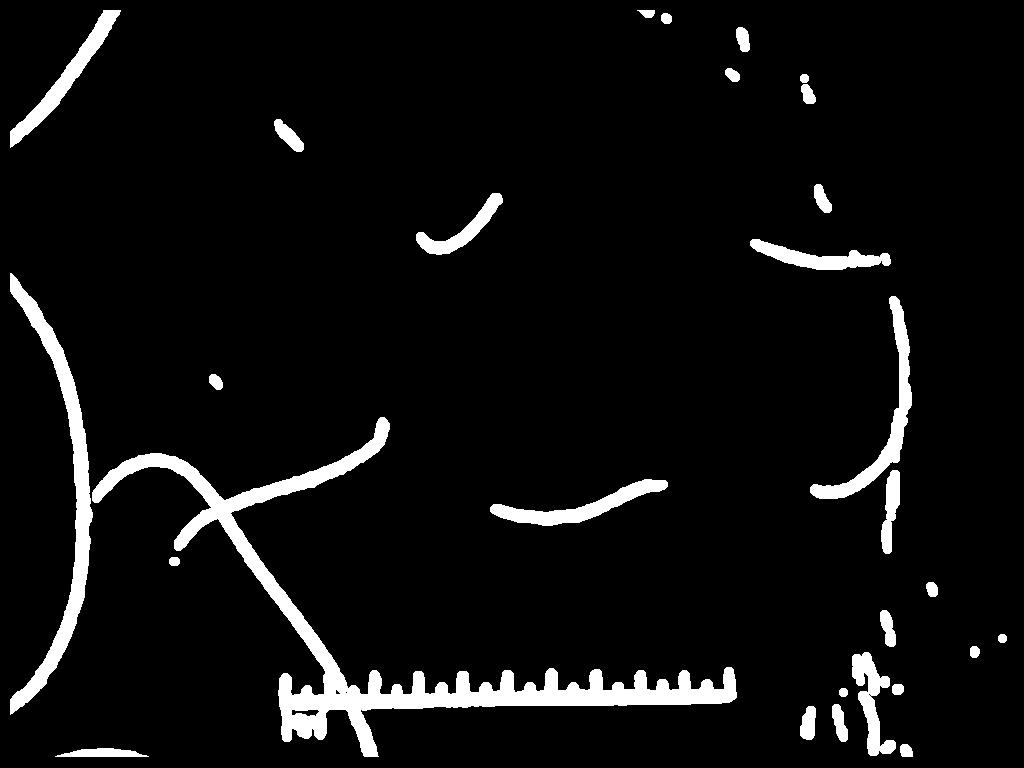}&\includegraphics[width=0.3\linewidth]{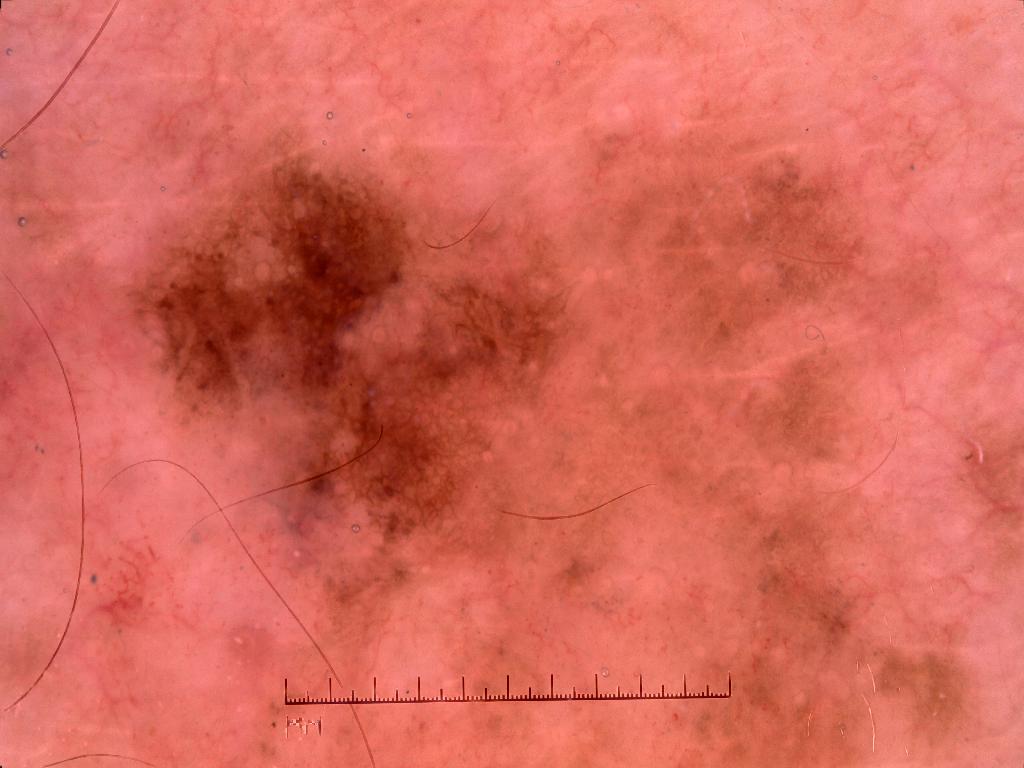}\\
    \includegraphics[width=0.3\linewidth]{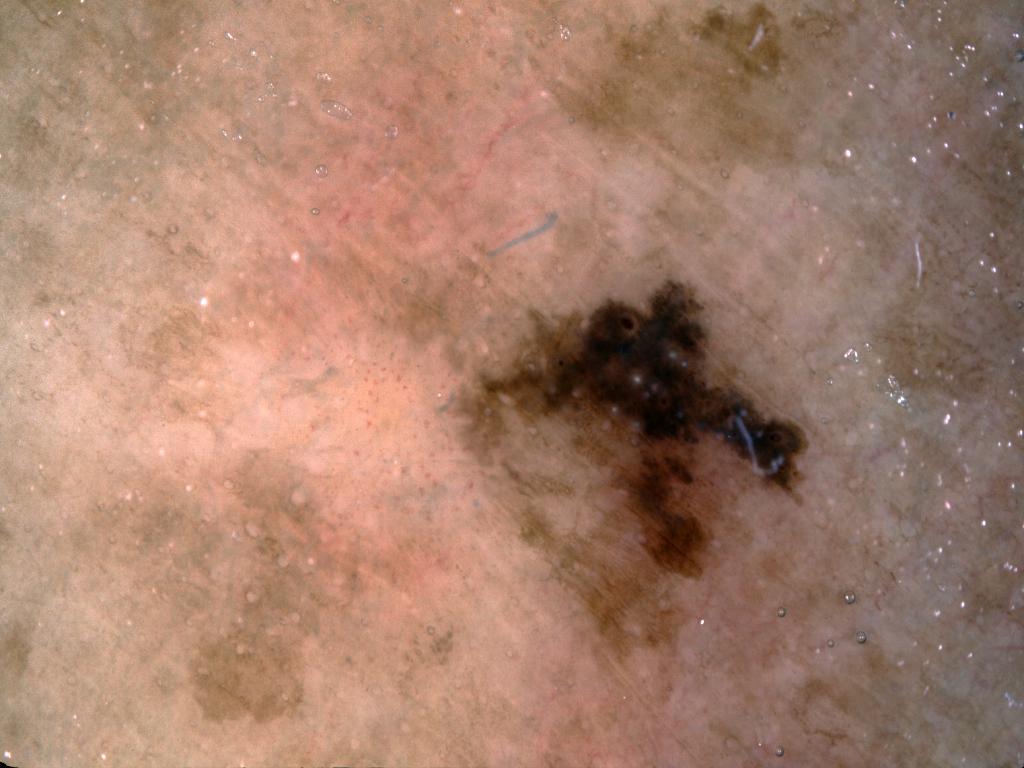}&\includegraphics[width=0.3\linewidth]{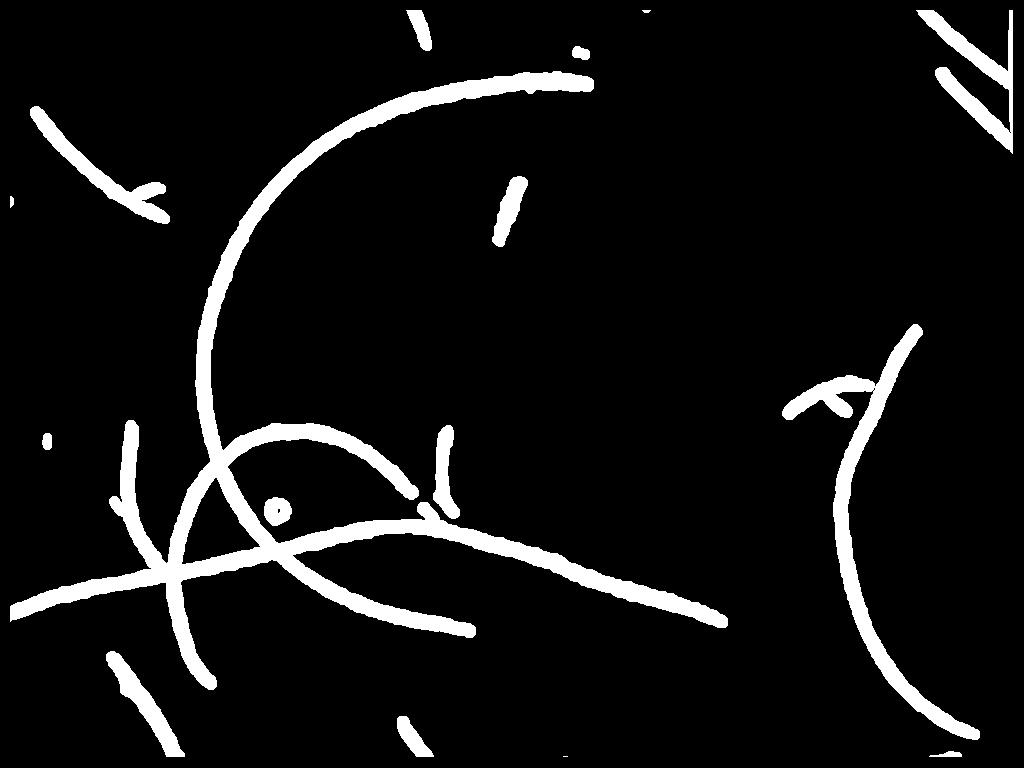}&\includegraphics[width=0.3\linewidth]{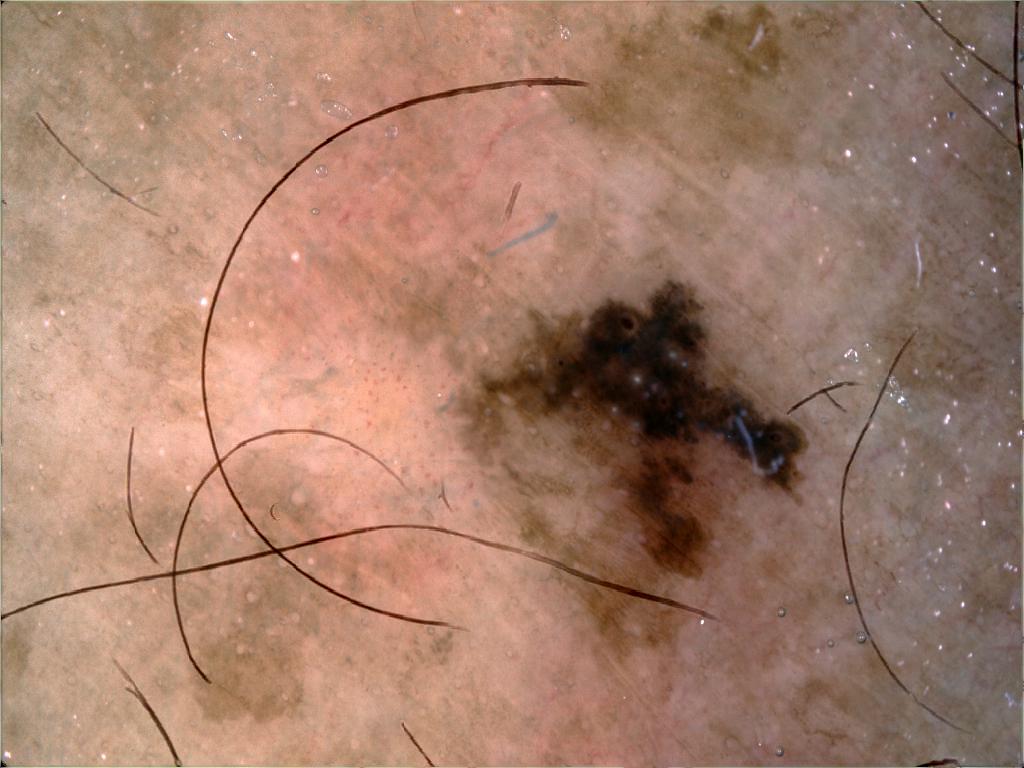}
    \end{tabular}
    \caption{Simulation results for the proposed method. The Hair has been synthesised according to the mask. Also, the proposed method is able to generate other artefacts such as ruler markers.}
    \label{fig:resultsl}
\end{figure*}

We prepared the hair masks using an ensemble of hair segmentation method. We used the proposed method by Koehoorn~\textit{et al.}~\cite{koehoorn2016effcient} and Zhou~\textit{et al.}~\cite{Zhou2013}. We used skin images from a publicly available dataset from ISIC archive~\cite{berseth2017isic} for validation and comparison purposes. This dataset consists of total 1279 images. We manually selected the hair-free images. We find a total of 72 hair-free images. We super imposed the segmented hair masks on them for simulation.
\smallbreak
For qualitative analysis, we super imposed random mask on hair-free images. We compared between the different images to study the effect of the hair colour on different skin lesions. As shown in Fig.~\ref{fig:resultsl}, the proposed method successfully blended the hair on the skin with more realistic appearance compared to the proposed method by Mirzaalian~\textit{et al.}~\cite{mirzaalian2014hair}. The proposed method was able to synthesis hair on different skin architecture. Also, it has been able to synthesis other image atrefacts such as ruler markers, as shown in Fig.~\ref{fig:resultsl}
\smallbreak
For quantitative analysis, we used a quantitative metric for blinded image quality in terms of realism called realismCNN~\cite{zhu2015learning}. This metrics is used to assess the compatibility of the colours of the colours inside the image to detected the realism of the objects in the images. We achieved realism score 1.65 based on realismCNN aganist 1.59 for the proposed method in~\cite{mirzaalian2014hair}.

\section{Conclusion}
\label{sec:CONCLUSION}
In this presented work, we proposed a novel methodology for realistic hair simulation. We utilised image blending techniques to obtain simulated hair, along with the corresponding mask. This proposed method can be used as a realistic hair simulator for data augmentation and for validation of hair segmentation and inpainting methods. We used state-of-the-art image blending technique to simulate hair on hair-free images. This method was able to simulate hair with matching colour to both source and destination. Thus, the output hair is realistic hair with high compatibility to the destination image. Also, the colours of simulated hair-masks have high degree of variability.


\section*{Acknowledgement}

This research was fully supported by the Institute for Intelligent Systems Research and Innovation (IISRI) at Deakin University.

\bibliographystyle{IEEEtran}

\bibliography{ref}

\end{document}